\documentclass[preprint,12pt,authoryear]{elsarticle}

\usepackage{amssymb}
\usepackage{amsmath}
\usepackage{enumerate}
\usepackage{graphicx}%
\usepackage{hyperref}
\hypersetup{hypertex=true,
	colorlinks=true,
	linkcolor=blue,
	anchorcolor=blue,
	citecolor=blue}
\usepackage{multirow}%
\usepackage{array}%
\usepackage{diagbox}%
\usepackage{epstopdf}%
\usepackage{booktabs}%
\usepackage{pdflscape}
\usepackage{manyfoot}%
\usepackage{listings}%
\usepackage{makecell}%
\usepackage{longtable}%
\usepackage{threeparttable}%
\usepackage{amsmath,amssymb,amsfonts}%
\usepackage{mathrsfs}%
\usepackage{xcolor}%
\usepackage{textcomp}%

\usepackage{algorithm}%
\usepackage{algorithmicx}%
\usepackage{algpseudocode}%
\usepackage{float}%
\usepackage{bm}
\usepackage[section]{placeins}

\begin{document}

\begin{frontmatter}

\makeatletter
\def\ps@pprintTitle{%
	\let\@oddfoot\@empty
	\let\@evenfoot\@empty
}
\makeatother

\title{SC4ANM: Identifying Optimal Section Combinations for Automated Novelty Prediction in Academic Papers}

\author{Wenqing Wu}
\ead{winchywwq@njust.edu.cn}

\author{Chengzhi Zhang \corref{cor1}}
\ead{zhangcz@njust.edu.cn}

\author{Tong Bao}
\ead{tbao@njust.edu.cn}

\author{Yi Zhao}
\ead{yizhao93@njust.edu.cn}

\cortext[cor1]{Corresponding author}

\affiliation{organization={Department of Information Management,Nanjing University of Science and Technology},            
            city={Nanjing},
            postcode={210094}, 
            country={China}}

\begin{abstract}
Novelty is a core component of academic papers, and there are multiple perspectives on the assessment of novelty. Existing methods often focus on word or entity combinations, which provide limited insights. The content related to a paper's novelty is typically distributed across different core sections, e.g., Introduction, Methodology and Results. Therefore, exploring the optimal combination of sections for evaluating the novelty of a paper is important for advancing automated novelty assessment. In this paper, we utilize different combinations of sections from academic papers as inputs to drive language models to predict novelty scores. We then analyze the results to determine the optimal section combinations for novelty score prediction. We first employ natural language processing techniques to identify the sectional structure of academic papers, categorizing them into introduction, methods, results, and discussion (IMRaD). Subsequently, we used different combinations of these sections (e.g., introduction and methods) as inputs for pretrained language models (PLMs) and large language models (LLMs), employing novelty scores provided by human expert reviewers as ground truth labels to obtain prediction results. The results indicate that using introduction, results and discussion is most appropriate for assessing the novelty of a paper, while the use of the entire text does not yield significant results. Furthermore, based on the results of the PLMs and LLMs, the introduction and results appear to be the most important section for the task of novelty score prediction. The code and dataset for this paper can be accessed at https://github.com/njust-winchy/SC4ANM.
\end{abstract}


\end{frontmatter}



\section{Introduction} \label{sec1}
\noindent Novelty is one of the core criteria in academic research, as it drives the frontier of knowledge, addresses unresolved questions in existing studies, or presents new insights. The current mainstream approach for evaluating the novelty of a paper utilizes the perspective of combinatorial innovation \citep{r1,r2,r3}, examining the distribution of references within the paper or the distribution of knowledge elements \citep{r4,r5}. However, although easy to understand and simple to use, the extent to which cited publications serve as sources of inspiration for an academic paper is not yet well understood \citep{r6}. In addition, currently most of the main content of academic papers is not well used to evaluate novelty, and most of them are evaluated in the form of knowledge entities, keywords or topic words in the abstract and title. Content or entities related to the novelty of a paper are typically distributed across various sections, and relying solely on the abstract or title cannot fully capture all the knowledge utilized in the paper. When evaluating the novelty of a paper, relying solely on entity or word-level content is limited. Furthermore, in previous novelty measures based on entities, the entire document was treated as a window for entities, without considering the impact of different section and their combinations on the novelty measurement. To address this limitation, it is essential to explore which sections or combinations of sections in a paper provide the most insightful information for an accurate automatic novelty evaluation.\\
\begin{figure*}[!h]%
	\centering
	\includegraphics[width=1\textwidth]{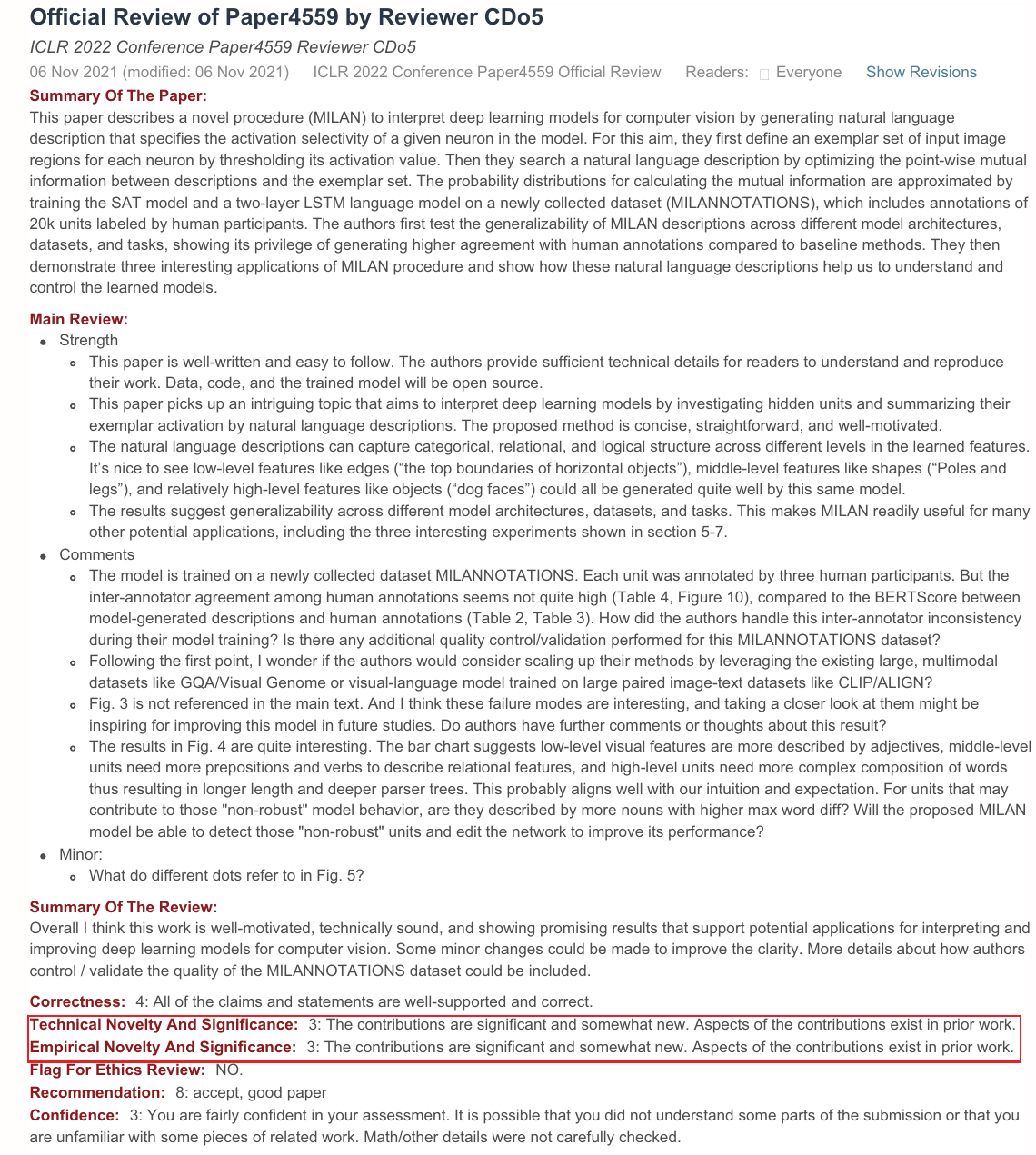}
	\vspace{-0.7cm}
	\caption{\centering{An Example of review report on ICLR 2022. (https://openreview.net/forum?id=ltM1RMZntpu)}}
	\label{fig:1}
\end{figure*}
\indent Generally, academic papers are divided into title, abstract, main text, and references. The main text can further be divided into introduction, methods, results, and discussion (IMRaD) \citep{r7}. Although not all academic papers follow this structure, natural language processing techniques \citep{r8} can be employed to segment the content of the main text into these sections. In the peer review process\footnote{https://www.nature.com/nature/for-referees/how-to-write-a-report. https://2023.aclweb.org/blog/review-acl23/. https://icml.cc/Conferences/2023/ReviewerTutorial. https://iclr.cc/Conferences/2024/ReviewerGuide. These reviewer report instructions clearly require reviewers to evaluate or grade the significance and novelty of the paper.}, the evaluation of an academic paper's novelty primarily relies on the judgment of expert reviewers. Reviewers typically do not base their assessment solely on the words or entities form the paper, but rather make their assessment after read the main text of the paper. Given that novelty often arises from the broader context and argumentation presented throughout the paper, it is crucial to understand how various sections contribute to the overall assessment. Consequently, simulating the review process by analyzing different combinations of paper sections can provide a more accurate prediction of a paper's novelty.\\
\indent With advancements in natural language processing, traditional machine learning and deep learning technology \citep{ref48,ref49}, particularly the emergence of large language models (LLMs) \citep{r10,r11}, the ability of machines to process academic papers has been significantly enhanced. LLMs have also demonstrated commendable performance in research related to peer review \citep{r12,r13}, although there remains a considerable gap compared to human reviewers. Current research using LLMs primarily focuses on the macro-level evaluation of academic papers. However, the assessment of the novelty of papers is often overlooked, lacking clear evaluation or validation.  \\
\indent With an increasing number of conferences, journals, and platforms such as NeurIPS\footnote{https://neurips.cc/Conferences/2022/CallForPapers}, eLife\footnote{https://reviewer.elifesciences.org/author-guide/editorial-process}, PeerJ\footnote{https://peerj.com/benefits/review-history-and-peer-review/}, OpenReview\footnote{https://openreview.net/about}, and F1000Research\footnote{https://f1000research.com/about} offering open peer review, the novelty scores provided by expert reviewers can be obtained from these publicly available peer review reports, as illustrated in the red box in Figure \ref{fig:1}. The novelty scores provided in the reviewers' reports can serve as the evaluation standard for a paper's novelty. Currently, a research \citep{r41} have been conducted using open peer review reports to analyze the consistency of text scores. We believe that the novelty scores given by reviewers can serve as a reference standard for evaluating the novelty of a paper.\\
\indent To identify and analyze the optimal sections or combinations of sections for evaluating the novelty of academic papers, we obtained all peer review reports and submitted paper PDFs from the ICLR 2022 and ICLR 2023 conferences on the OpenReview platform. These peer review reports require reviewers to provide novelty scores for the papers. Based on this dataset, we conducted an empirical study on predicting the novelty of papers using different combinations of academic paper sections. Additionally, we validated the performance of LLMs on this task.  \\
\indent The study is driven by the following three research questions.\\
\indent \textbf{RQ1}: Which combinations of sections yield novelty scores that most closely align with the ground truth scores? By identifying which combinations of sections yield prediction scores closest to the actual novelty scores, we can assess the model's accuracy. This means that, based on these combinations, the model can predict results that more closely align with the true novelty scores of a paper, thereby reducing bias and errors. Additionally, this approach helps validate the model's effectiveness under different combinations, revealing which configurations best replicate human reviewer judgments, thus ensuring the model's rationality and credibility.\\
\indent \textbf{RQ2}: Which combinations of sections are most effective in predicting the novelty scores of academic papers? Different sections of a paper (such as Introduction, Methods, Results) may have varying impacts on novelty scores. Understanding how these sections can be combined to produce the most accurate predictions not only aids in improving novelty scoring algorithms but also provides deeper insights into the relationship between paper structure and content.\\
\indent \textbf{RQ3}: Which combinations of sections should be prioritized when automatically evaluating the novelty of a paper? For large-scale automated review systems, it is crucial to prioritize which sections of a paper should be focused on during the automatic evaluation process, as this can enhance both the efficiency and accuracy of the review. By prioritizing the sections most representative of novelty, the influence of subjective human factors can be minimized, making the evaluation more objective and fairer.\\
\indent The main contributions of this paper are reflected in the following three aspects.\\
\indent Firstly, we used the novelty scores from peer reviews as a benchmark to fine-tune current popular pre-trained language models (PLMs) designed for long texts to predict novelty scores. We also validated the effectiveness of using different section structures as input.  \\
\indent Secondly, we conducted a small-scale test on novelty score prediction using prompt-based methods on LLMs. Furthermore, we analyzed the performance of the LLMs in this task under different chapter combinations, as well as the consistency between generated novelty scores and grounded scores.\\
\indent Thirdly, the results indicate that fine-tuned PLMs outperform LLMs in predicting novelty scores, though their performance is not yet satisfactory. Furthermore, our findings suggest that the introduction, results, and discussion sections are more beneficial for automatic novelty score prediction tasks.\\
\indent All the data and source code of this paper are freely available at the GitHub website: https://github.com/njust-winchy/SC4ANM.

\section{Related work}
\label{sec2}
\noindent In this section, we will report related work about our research. Firstly, we present relevant work in the field of identification of academic paper structure. Subsequently, we delve into literature concerning LLMs for reviewing, followed by an overview of studies investigating novelty assessment methods.

\subsection{Novelty and other relevant concepts}
To formally define novelty, we first clarify the distinctions between novelty and other related concepts, such as innovation, disruptiveness, and originality.\\
\indent In the academic community, novelty is currently defined in several ways. One definition \citep{ref50} defines novelty as the degree of difference between a new scientific article and the existing body of scientific literature. Another definition \citep{ref51} defines novelty as the uniqueness of specific knowledge elements, where the inclusion of new knowledge elements in a scientific article indicates that it conveys novel information. In addition, there is another definition \citep{ref52} considers novelty as the result of a new combination of knowledge elements. Overall, the novelty of scientific articles can be unified as the quality of presenting new information within these articles.\\
\indent Innovation \citep{ref53} can be defined as the transformation of new ideas into practical, valuable, or sustainable outcomes. Unlike mere novelty, innovation must also generate impact in practice. Novelty is merely the first step toward innovation \citep{ref54}, which further requires the realization of practicality, impact, and value. Novelty can emerge in various settings, such as universities, whereas innovation primarily takes place within firms operating in the commercial sector \citep{ref55}.\\
\indent Disruptiveness \citep{ref56} refers to research or discoveries that fundamentally alter existing academic paradigms, research methods, or core theories within a field. This change not only impacts academic discourse but may also influence societal, industrial, and even policy-level transformations. Disruptiveness \citep{ref57} is often driven by research outcomes that break conventional norms and propose revolutionary new ideas. In contrast to novelty, disruptiveness not only emphasizes the origins of scientific research but also considers its utility and impact \citep{ref58}.\\
\indent Originality \citep{ref59,ref60} can be defined as the generation of new ideas, methods, conclusions, and other valuable outputs that depart from existing knowledge, or as a catalyst for further innovation. In practical measurement, distinguishing between originality and novelty is often challenging, as originality is typically implicit in research papers \citep{ref61}. Consequently, in most cases, originality and novelty are frequently used interchangeably \citep{r3,ref59,ref57}.\\
\indent In summary, novelty forms the foundation of both innovation and disruptiveness and is synonymous with originality. Without novel ideas or discoveries, it is impossible to develop innovative or disruptive outcomes. While scholars may offer varying interpretations of novelty, there is a consensus that in scientific papers, novelty refers to the intrinsic quality of presenting new knowledge. Specifically, a novel scientific paper addresses new research questions, methodologies, results, theories, or the reorganization of existing knowledge elements. Therefore, in this paper, we explore the prediction of novelty scores from the perspective of different combinations of sections. Specifically, various sections of a scientific paper, such as the Introduction, Methods, Results, and Discussion, often carry different types of information, which play a crucial role in presenting novelty.

\subsection{Novelty measurements of scientific publications}
\label{subsec2.3}
\noindent Novelty is one of the key criteria for evaluating the quality of academic papers. In the current community of scholars, novelty is defined as the reorganization of existing knowledge components in an unprecedented manner \citep{r30,r31}. Currently, scholars assess the novelty of a study by evaluating the new combinations of references, considering that knowledge reorganization is based on the content of the references section. Uzzi et al. \citep{r3} analyzed 17.9 million papers on Web of Science to study the relationship between reference combinations and citation counts. They found that the most impactful papers combine typical earlier works with new, unique combinations. This discovery helps distinguish between novel and less novel papers. Wang et al. \citep{r2} explored the relationship between the novelty of scientific research, defined by first-time combinations of referenced journals, and its long-term impact. They find that highly novel research offers significant benefits but also involves higher risks, often faces delayed recognition, and is published in lower Impact Factor journals. Matsumoto et al. \citep{r1} applied a novelty indicator to quantify the reference similarity between a focal paper and pre-existing papers within the same domain across various fields of natural sciences. They proposed a new method for identifying papers belonging to the same domain as the focal paper using only bibliometric data. Shibayama et al. \citep{r32} developed a more integrated method that utilizes both references and the content of a paper. In their approach, they quantify the semantic distance of references to assess the novelty of a given paper. \\
\indent However, although the reference is easy to understand and use, it remains unclear to what extent they can serve as a source of inspiration for academic papers \citep{r6}. Additionally, references in papers are provided solely by the authors themselves, without any oversight mechanism to ensure their quality. Recently, methods for discovering innovative content based on text analysis have been gaining popularity. Luo et al. \citep{r5} introduced a novel approach to measuring the novelty of papers from the perspective of problem method combinations. They proposed a semantic novelty measurement algorithm based on term semantic similarity, and evaluated the effectiveness of the method through case studies and statistical analysis. Yin et al. \citep{r33} developed a word embedding model using machine learning to extract semantic information related to elements of knowledge innovation from textual data. Jeon et al. \citep{r4} proposed an analytical framework that uses paper titles to measure the novelty of scientific publications. Chen et al. \citep{r42} conducted an in-depth investigation into the relationship between the institutional composition of author teams and the novelty of academic papers, using fine-grained knowledge entities to measure the novelty of the papers. Through case studies, they demonstrated that this framework is a useful supplementary tool. Liu et al. \citep{r47} explored the pursuit of scientific novelty in doctoral dissertations by Ph.D. students and evaluated the gender-related differences in this process. Wang et al. \citep{r9} proposed MNSA-ITMCM framework integrates topic modeling and a cloud model to measure novelty in scientific articles by combining semantically informed topics and quantifying novelty to improve accuracy, demonstrated through empirical evaluations in biomedical and computer science domains.\\
\indent The aforementioned research methods have demonstrated their effectiveness to some extent, indicating that text analysis-based approaches can be used to evaluate novelty. However, these methods only utilized parts of the paper, such as titles, abstracts, or knowledge entities, for their assessments and did not consider the section content of the paper. The novelty of a paper cannot be accurately determined based on brief excerpts alone. During the peer review process, reviewers typically assess the novelty of a paper by reading its entire content or most of its sections. Therefore, it is crucial to explore which sections should be given priority when evaluating the novelty of a paper. This paper simulates human reading of different sections using artificial intelligence and uses the novelty scores provided by reviewers as a benchmark to investigate the importance of various sectional content combinations.
\subsection{Large Language Models for Reviewing}
\label{subsec2.2}
\noindent With the rapid advancement of artificial intelligence and natural language processing technologies, LLMs \citep{r21,r22}, particularly those supported by transformer-based architectures and pre-trained on massive datasets, have gradually come into the public eye. With the successive releases of ChatGPT and GPT-4 \citep{r10} by OpenAI, as well as Llama \citep{r23} and others \citep{r11,r24,r25}, these LLMs have demonstrated powerful language generation and understanding capabilities, attracting significant research interest from scholars in the academic community. \citep{r43,r45,r46} \\
\indent Recently, research on using LLMs  for peer review has been gaining increasing popularity. Liang et al. \citep{r12} conducted a large-scale empirical analysis to assess comments generated by GPT-4: they tag several metrics and investigated user satisfaction to measure comment quality. They found that while LLM-generated reviews were helpful, they might be non-generic and tended to focus on certain aspects of scientific feedback. Liu and Shah \citep{r26} validated the utility of LLMs across three tasks: identifying errors, verifying checklists, and choosing the "better" paper. They concluded that LLMs serve well as review assistants for specific reviewing tasks; however, they are not yet sufficient for conducting comprehensive evaluations of papers. Mike Thelwall \citep{r27} used GPT-4 to evaluate the quality of journal articles on a paper assessment dataset to test its effectiveness. The results indicated that GPT-4 appears to be insufficiently accurate for any formal or informal research quality assessment tasks. Zhou et al. \citep{r13} conducted a comprehensive evaluation to determine whether LLMs can be qualified and reliable reviewers. They concluded that it is premature for LLMs to serve as automated scientific paper reviewers. Although there is potential for obtaining useful and accurate results, their current capabilities are not yet reliable enough. Robertson \citep{r28} conducted a preliminary study on using GPT-4 to assist in the peer review process, providing initial evidence that artificial intelligence can effectively facilitate this process. Gao et al. \citep{r29} proposed an efficient two-stage review generation framework REVIEWER2 that simulates the distribution of potential aspects a review might address. Additionally, they generated a large-scale peer review dataset comprising 27,000 papers and 99,000 reviews based on this framework.\\
\indent However, these studies are all macro-level investigations, focusing on the overall evaluation of academic papers. Our research, in contrast, focuses specifically on the novelty of academic papers and examines the ability of LLMs to assess the novelty of academic papers.\\
\subsection{Identification of Academic Paper Structure}
\label{subsec2.1}
\noindent The IMRaD model is a classical system, proposed earlier and widely used in scientific literature \citep{r7,r14}. It divides the structural functions of academic articles into four parts: Introduction, Methods, Results, and Discussion. The purpose of identifying the sectional structure of academic papers is to determine their structural functions, specifically, to classify text segments (sentences, paragraphs, or sections) into their respective functional categories. Lu et al. \citep{r15} proposed a clustering method based on domain-specific structures using high-frequency section headings in scientific documents to automatically identify the structure of scientific literature. They applied the proposed method to two tasks: academic search and keyword extraction, achieving good performance. Li et al. \citep{r16} proposed a hybrid model that considers both section headings and the main text to automatically identify general sections in academic literature. Ma et al. \citep{r17} utilized section headings to identify the structural functions of academic articles, incorporating relative position information and contextual information.\\
\indent The aforementioned studies have made significant progress in the task of sectional structure identification. Subsequent research on this topic has generally included relevant applications. Cohan et al. \citep{r18} proposed a novel hierarchical encoder for modeling the discourse structure of documents, applied across two large scientific paper datasets. Ji et al. \citep{r19} proposed using deep learning to automatically identify the functional structure of academic texts based on section content, laying the groundwork for research on the distribution of reference locations. Qin et al. \citep{r20} explored which sections of academic articles reviewers focus on most by analyzing the sectional structure, as well as identifying the specific content that reviewers pay attention to. Although the above research has done a lot of work on identifying the structure of academic papers, we believe that identifying the structure of academic papers is still a challenging task. Zhou and Li \citep{r44} investigated the problem of section identification in academic papers within the context of Chinese medical literature. They employed effective features from classical machine learning algorithms to address this issue.\\
\indent In our approach, we use deep learning models and LLMs to identify the sectional structure, and we perform consistency checks between their results to ensure the accuracy of the identification process.

\section{Methodology}
\label{sec3}
\noindent In this section, we will introduce our data collection and preprocessing process, as well as the novelty score prediction task. Subsequently, we will provide a detailed explanation of the methods applied for section structure identification and novelty score prediction.\\
\definecolor{blue5}{rgb}{0.2, 0.4, 0.8}
\definecolor{green6}{rgb}{0.4, 0.8, 0.4}
\definecolor{orange2}{rgb}{1.0, 0.6, 0.2}
\begin{figure*}[h]%
	\centering
	\includegraphics[width=1\textwidth]{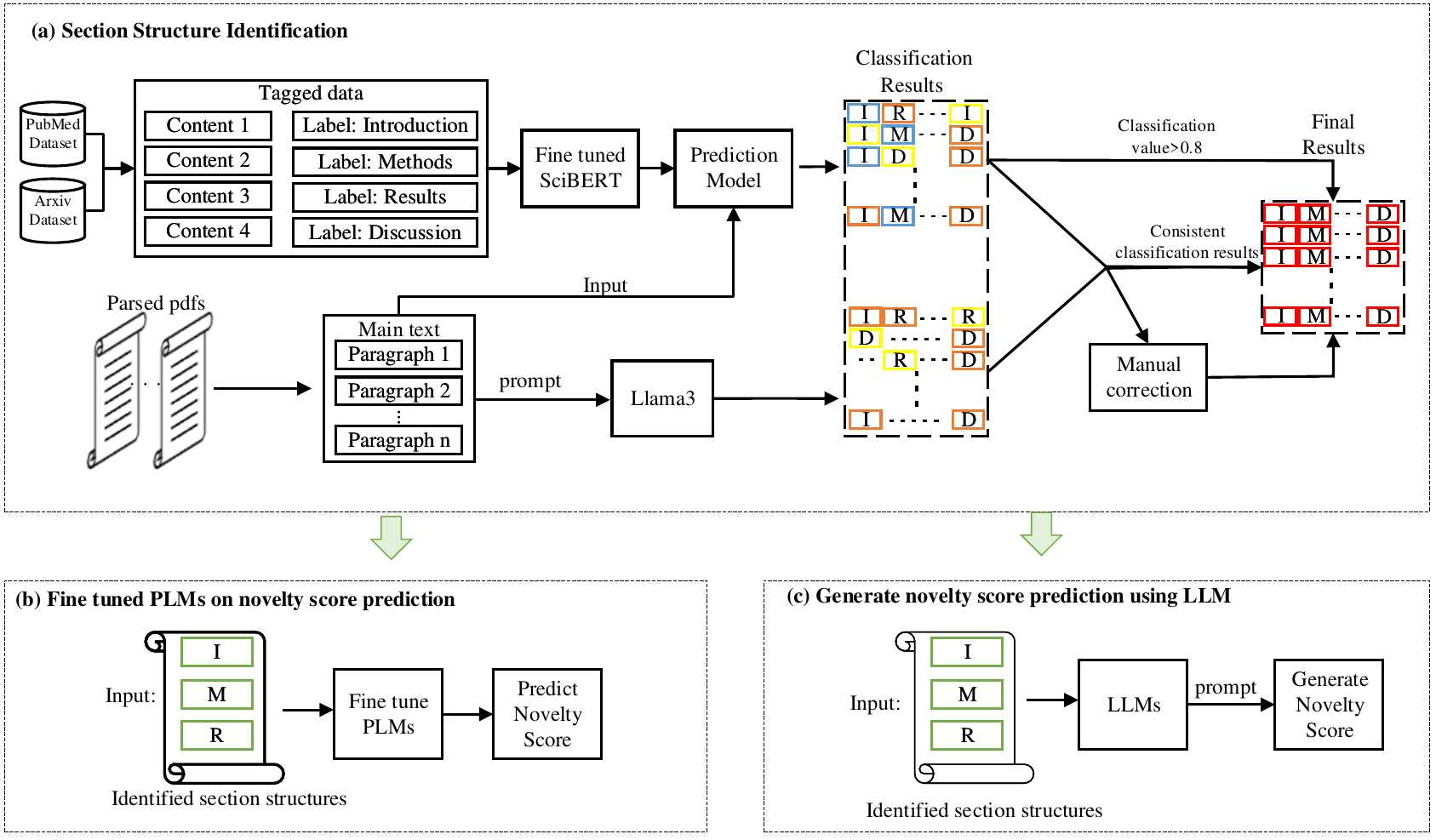}
	\vspace{-0.9cm}
	\begin{flushleft}
		{\footnotesize \textbf{Note}: The combination of section structure takes the introduction (I), methods (M), and results (R) as an example. The green boxes represents the text of the corresponding section that has been identified. The \textcolor{blue5}{blue boxes} indicate that the PLM's predicted classification value for the text is greater than 0.8. The \textcolor{orange2}{orange boxes} represent cases where the PLM's classification result is less than 0.8, but the LLM's prediction is consistent with it. The \textcolor{yellow}{yellow boxs} indicate that PLM's predicted classification value for the text is less than 0.8 and inconsistent with the LLM, require human correction. \textcolor{red}{Red boxs} are the final result.} 
	\end{flushleft}
	\caption{\centering{Framework of this study.}}
	\label{fig:2}
\end{figure*}
\indent The following is the research method of this paper. Figure \ref{fig:2} provides an overview of the methodology used in this study,  which includes Section  Structure Identification, Fine tuning PLMs  for novelty score prediction, and Generate novelty score prediction using LLM.
\subsection{Data collection and preprocessing}
\label{subsec3.1}
\noindent We obtained our peer review data from the OpenReview platform. The International Conference on Learning Representations (ICLR) is a premier conference in the field of machine learning. We wrote a web crawler code to retrieve a total of 8183 ICLR papers from ICLR 2022 and 2023, each containing peer review comments. The reason for selecting papers from these two years as the data source is that the review reports for these years require reviewers to provide novelty scores, as shown in Figure \ref{fig:1}. Additionally, another reason for selecting data from these two years as the source for this study is that the common section types in ICLR papers closely align with the IMRD structure typically found in natural sciences. In the data we collected, all ICLR papers include an introduction, the section corresponding to the methods is generally labeled as "Methodology" or the specific name of the proposed method, the section corresponding to the results is usually related to experiments or model comparisons, and the section corresponding to the discussion typically includes analysis, discussion, and conclusions. Therefore, we can map these sections to the corresponding IMRD structure using a section structure recognition method in Section \ref{subsec3.3}. It is also worth noting that the methods proposed in this paper are applicable in any domain where the sections of a paper can be classified.\\
\indent The peer review reports for ICLR 2022 and 2023 included the review text, Correctness, Technical Novelty and Significance, Empirical Novelty and Significance, Recommendation, and Confidence. Among these, the distribution of Technical Novelty and Significance (TNS) scores ranges from 1 to 4, denoted as 1: The contributions are neither significant nor novel, 2: The contributions are only marginally significant or novel, 3: The contributions are significant and somewhat new. Aspects of the contributions exist in prior work and 4: The contributions are significant, and do not exist in prior works, respectively. As ICLR is a conference in the field of computer science, we have chosen TNS  as the criterion for predicting novelty scores. Each paper receives evaluations from at least three reviewers, and each reviewer assigns a TNS score. To obtain labels suitable for model training, we aggregate the scores provided by each reviewer by summing them up and then dividing by the total number of reviewers. However, discrepancies may arise when multiple reviewers have differing opinions, such as one assigning a score of 4 and another a score of 1. To address this issue, we identified the maximum and minimum scores in each review report and exclude instances where the difference exceeds 1, signifying reviewer disagreement. These cases are subsequently removed from the dataset. After the consistency processing of the scores mentioned above, there are ultimately 6,094 papers with peer review reports that have consistent TNS scores. \\
\begin{table}[h]
	\caption{\centering{Statistical results of the novelty score data.}}\label{tab1}%
	\centering
	\begin{tabular}{@{}m{40mm}<{\centering} m{28mm}<{\centering}m{28mm}<{\centering}m{28mm}<{\centering} @{}}
		\toprule
		\diagbox{\textbf{TNS}}{\textbf{Decision}} & \textbf{Accept} & \textbf{Reject} & \textbf{Total} \\
		\midrule
		\# TNS=1    & 0 & 60 & 60 \\
		\# TNS=2    & 272 & 1726 & 1998\\
		\# TNS=3    & 750 & 599 & 1349 \\
		\# TNS=4    & 87 & 6 & 93  \\
		\# Papers  & 1109  & 2391 & 3500\\
		\bottomrule 
	\end{tabular}
	\begin{tablenotes}
		\footnotesize
		\item \textbf{Note:} TNS is Technical Novelty and Significance score.
	\end{tablenotes}
\end{table}
\indent Furthermore, to parse the acquired academic paper PDFs, we utilized the GROBID (2008-2022)\footnote{https://github.com/kermitt2/grobid} and S2ORC \citep{r34} tools to parse the PDFs into JSON format. In the large-scale dataset constructed by Gao et al. \citep{r29}, the parsed content data of papers from ICLR 2022 and 2023 is included, eliminating the need for custom rules to extract paper content from JSON files. We matched the parsed paper abstracts with the dataset created by Gao et al. \citep{r29} and ultimately matched the content of 3,500 papers. We investigated the reason for the significant reduction in the number of papers and found that many PDF files were non-editable, meaning they could not be parsed, or the matching process failed. The final distribution of novelty scores is shown in Table \ref{tab1}. From the Table \ref{tab1}, it is evident that the counts for scores 1 and 4 are relatively low. Predicting directly with such an extremely imbalanced label distribution may result in biased prediction outcomes. Therefore, to address this issue and considering the importance of a novelty score of 4, we reclassified the scores by combining scores 1 and 2, retaining score 3 and score 4 as a separate category. According to the descriptions associated with each score, the novelty of paper is classified into three categories: 0: basic novel, 1: moderate novel, 2: highly novel.
\subsection{Problem definition}
\label{subsec3.2}
\noindent In this work, we introduce a new task called optimal section combinations for automated novelty measurement (SC4ANM). The objective of SC4ANM is to predict the novelty scores of academic papers based on varying combinations of their sections. The task aims to evaluate how various combinations of a paper's sections impact to its novelty score. Formally, given an academic paper, let $P=(S_1,S_2,...,S_n)$ represent the set of sections of the paper, where each section corresponds to a part of the paper (e.g., Introduction, Methods, Results, Discussion, etc.). A section combination $C$ is defined as a subset of these sections, $C\subseteq P$ selected for analysis. For example, combinations such as Introduction and Methods, Introduction and Results, or even Introduction only are valid section combinations. A set of corresponding labels $L=\{l_0:0,l_1:1,l_2:2\}$ represents the novelty scores defined in the end of Section \ref{subsec3.1}. The goal of SC4ANM is to develop a classification model $f$ that assigns predefined novelty scores $l_{\{0,1,2\}}$ based on the content of different section combinations selected.
\subsection{Section Structure Identification}
\label{subsec3.3}
Firstly, we need to identify the structure of the parsed paper, dividing the main text into introduction, methods, results, and discussion.\\
\indent However, not all academic papers adhere to the IMRaD structure, thus necessitating the identification of section structures in the parsed PDF content. The parsed paper data not only includes the main text of the papers but also contains some additional content, such as page number and page header, quotes or excerpts etc. Therefore, it is necessary to identify the main text of the papers to remove any extraneous information. We utilized an academic text classifier, fine-tuned on linguistic academic publications\footnote{https://huggingface.co/howanching-clara/classifier\_for\_academic\_texts}, as the identification model to recognize the main content in our paper parsed data. We saved the extracted main text of academic papers in the form of paragraphs or sentences.\\
\begin{figure*}[h]%
	\centering
	\includegraphics[width=1\textwidth]{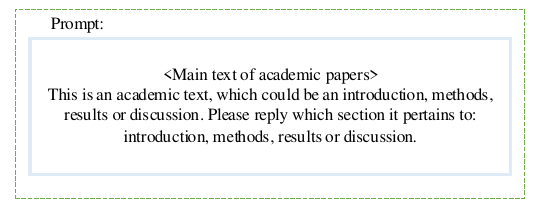}
	\vspace{-0.9cm}
	\caption{\centering{Prompt of Llama 3 for section structure identification.}}
	\label{fig:3}
\end{figure*}
\indent Then, we perform section structure identification on the extracted main text. Cohan et al. \citep{r18} provided two datasets of scientific papers. The datasets are sourced from the ArXiv and PubMed OpenAccess repositories, both of which provide the names of sections and their corresponding content. We extracted all IMRaD section content from the two datasets , totaling 300,000 entries, and divided them into training, validation, and test sets in an 8:1:1 ratio. We then fine-tuned the SciBERT \citep{r35} on this data to train a four-class classifier aimed at determining which section structure a given piece of academic text belongs to. The final model achieved a classification accuracy rate of 91\%. We believe that relying solely on the fine-tuned PLM does not achieve the best recognition results. Therefore, we also used LLaMA3 \citep{r11} for secondary recognition of the extracted main text. The specific prompt is shown in Figure \ref{fig:3}, the content within angle brackets represents the main text of the paper that require section structure recognition. If the predicted value for a given class from the PLM exceeds 0.8, we consider the prediction to be correct. Otherwise, further validation of the classification result is required. The threshold is set to 0.8 because we consider section structure identification to be a crucial task that should prioritize minimizing false positives. In other words, it is preferable to conduct further validation rather than risk incorrectly classifying sections as correct. Based on this application requirement, we believe that setting the threshold to 0.8 is a reasonable choice. We employed LLama3 for secondary validation. When the predictions from the trained model are consistent with the results from LLaMA3, we consider the prediction to be correct. When the predictions are inconsistent, we save the text and manually identify the section structure it belongs to. At this point, we have obtained the identified section structures and their corresponding content for each paper.
\subsection{Fine tuning PLMs for novelty score prediction}
\label{subsec3.4}
\begin{figure*}[h]%
	\centering
	\includegraphics[width=1\textwidth]{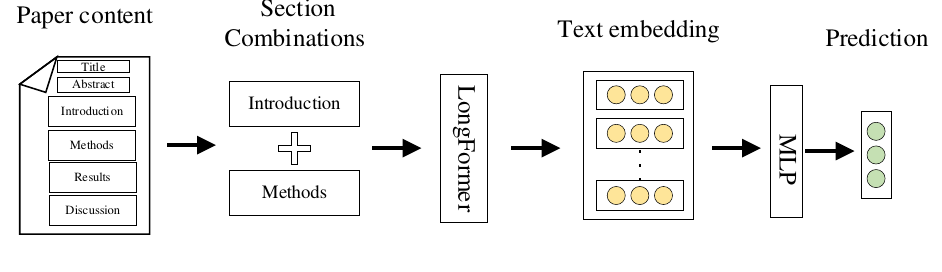}
	\vspace{-0.9cm}
	\caption{\centering{An Example of Fine-tuning PLMs for Novelty Score Prediction. Using Longformer to represent the PLM and Introduction + Methods to represent the section structure combinations as examples.}}
	\label{fig:4}
\end{figure*}
\noindent We fine-tuned five PLMs  designed for handling long texts to accomplish the task of novelty score prediction. These models include Longformer \citep{r36}, BigBird \citep{r37}, LongT5 \citep{r38}, LED (Longformer Encoder-Decoder) \citep{r39}, and the Longformer version of SciBERT\footnote{https://huggingface.co/yorko/scibert\_scivocab\_uncased\_long\_4096}. We divided the academic paper data, with identified section structures, into training, validation, and test sets in an 8:1:1 ratio, with 2500, 350, and 350, respectively. Next, as illustrated in the lower left corner of Figure \ref{fig:2}, we input different combinations of section structures (such as “Title+Abstract”, “Introduction”, “Introduction+Methods”, etc., totaling 16 combinations) to fine-tuned the PLMs. The results are then fed into an MLP to obtain the final prediction, which is the three-class classification of the novelty score. The detailed process is shown in Figure \ref{fig:4}.
\subsection{Generate novelty score prediction using LLM}
\label{subsec3.5}
\begin{figure*}[h]%
	\centering
	\includegraphics[width=1\textwidth]{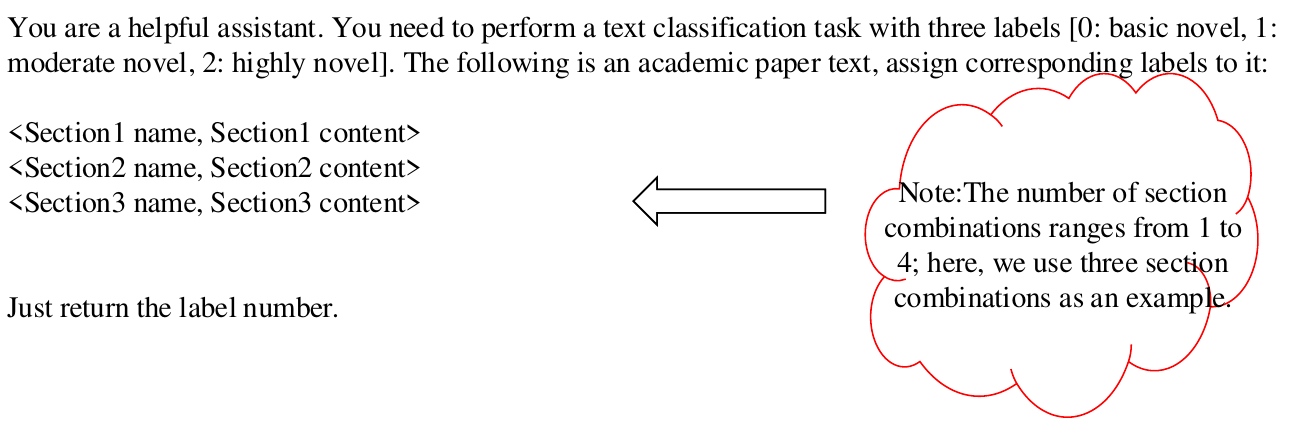}
	\vspace{-0.9cm}
	\caption{\centering{An Example used to prompt the large language model for the novelty score prediction task. }}
	\label{fig:5}
\end{figure*}
\noindent In addition to fine-tuning PLMs for novelty score prediction, we also tested the performance of LLMs  in predicting novelty scores under different section structure combinations. From the data statistics in Table \ref{tab1}, it can be observed that the number of papers with a novelty score of 4 is relatively low. Therefore, we randomly selected 40 papers with a novelty score of 4 as the experimental subjects. Additionally, the number of papers with other scores was made equal to that of the score 4 papers, also selected randomly. Notably, to ensure the diversity of experimental data, we performed random sampling of the papers again for each different combination of section structures used as input. For the selection of LLMs, we chose GPT-3.5 and GPT-4o as the models for our experiments. By calling their APIs and using prompt-based learning, we obtained predictions for the novelty scores of academic papers. The specific prompt is shown in Figure \ref{fig:5}. It is important to note that although we instructed the model to return only numerical classification labels in the prompt, GPT-3.5 still generated some additional content. Consequently, we performed a secondary verification of GPT-3.5's outputs. GPT-4o did not exhibit this issue.
\section{Result analysis}
\label{sec4}
\noindent In this section, we present the results of predicting novelty scores using different section structures and explore the following three research questions based on these results. 
\subsection{Evaluation Metrics on the SC4ANM task.}
\label{sec4.1}
\noindent The Accuracy (Acc) and $F{_1}$-score were selected to evaluate classification performance. The formulas used are as follows:
\begin{equation}
	Accuracy=\frac{ {\textstyle} \sum TP_i} {\text{The total number of sample}} 
\end{equation}
 Where $TP_i$ is the number of correct predictions for category $i$ (0: basic novel, 1: moderate novel, 2: highly novel). The calculation method for the $F{_1}$ score of each category $i$ is as follows:
 
 \begin{equation}
 	Precision_i= \frac{TP_i}{TP_i+FP_i}  
 \end{equation}
 \begin{equation}
 	Recall_i = \frac{TP_i}{TP_i+FN_i}  
 \end{equation}
\begin{equation}
	F{_1}_i=2\times \frac{Precision_i\times Recall_i}{Precision_i+Recall_i} 
\end{equation}
Where $i$ represents the category index (0, 1, 2), $TP_i$ is the number of correct predictions for category $i$, $FP_i$ is the number of incorrectly predicted as category $i$ but does not actually belong to class $i$, $FN_i$ represents the number of instances that actually belong to category $i$ but are incorrectly predicted as another category.\\ 
\indent Since the weighted average $F_1$ score is calculated by weighting each category according to its sample size, it is more suitable for imbalanced datasets. Therefore, we utilize the weighted average $F{_1}$ score, the weight for each category $i$:

\begin{equation}
	Weighted\_F{_1}=\sum \frac{c_i}{c} \times F{_1}_i
\end{equation}
Where $c_i$ represents the number of instances for category $i$, $c$ denotes the total number of samples, $F{_1}_i$ is the $F_1$ score for category $i$.\\
\indent In addition to utilizing common metrics for evaluating classification performance, we also employed the Pearson, Spearman and Kendall’s tau correlation coefficients, which are typically used to assess the score prediction capability of models. The formula for calculating the Pearson correlation coefficient is as follows:

\begin{equation}
	r=\frac{Cov(x,y)}{\sigma_x \sigma_y}  
\end{equation}
Where $Cov$ is covariance, $\sigma$ is standard deviation. The Spearman formula is applied:

\begin{equation}
	r_s=1-\frac{6n\overline{d^2}}{n^2-1} 
\end{equation}
\begin{equation}
	\overline{d^2}=\frac{1}{n} \sum_{i=1}^{n} d_i^2 
\end{equation}
\begin{equation}
	d_i=r_x(i)-r_y(i) 
\end{equation}
Where $r_x(i)$ and $r_y(i)$ represent the rank of $x_i$ and $y_i$, $n$ is the total number of samples. The formula for Kendall's Tau correlation coefficient is as follows:
\begin{equation}
	\tau =\frac{2(C-D)}{n(n-1)} 
\end{equation}
Where $C$ represents the number of concordant category pairs (i.e., when $x_i>x_j$ and $y_i>y_j$, or $x_i<x_j$ and $y_i<y_j$), $D$ represents the number of discordant category pairs (i.e., when $x_i>x_j$ and $y_i<y_j$, or $x_i<x_j$ and $y_i>y_j$), $n$ is the total number of samples.
\subsection{Traditional novelty measurement methods as baselines}\label{sec4.2}

\noindent We chose the novelty measurement method proposed by Shibayama et al. \citep{r32} as the representative of traditional approaches for comparison as the baseline. Their method calculates the novelty score by determining the semantic distance between the word vectors of the paper's title and those of the references' titles. We then simply present their novelty calculation method. First to calculate the distance between each pair of cited documents, the cosine distance between $i-th$ and $j-th$ references$(1\le i\le j\le N)$ is given by:
\begin{equation}
	d_{ij}=1-\frac{v_i\cdot v_j}{\left | v_i \right |\left | v_j \right | }  
\end{equation}

Where $v$ is mean of word embeddings of all words included, $N$ is number of the reference. Then calculate the distance scores($Novel_q$):
\begin{equation}
		Novel_q=R^{-1}\left ( \frac{N_q}{100}  \right )   
\end{equation}
	
Where $R(d_{ij})$is the ordinal rank of $d_{ij}$ of all the distances of $N(N-1)/2$ reference pairs, $q$ represents the percentile rank of the distance score, with a range of [0,100]. According to the findings in Shibayama et al.'s\citep{r32} study, setting $q$ to 100 yields the best results; therefore, we also set $q$ to 100. Since the computed results $Novel_q$ fall within the range [0,1], we apply a linear scaling method to map the results to the range [0,2] and round them to the nearest integer:
\begin{equation}
		scaled_{value}=2\times \frac{Novel_q-min(value_{list})}{max(value_{list})-min(value_{list})}    
\end{equation}

Where $value_{list}$ is the list of all calculated distance scores $Novel_q$, and $min(value_{list})$ and $max(value_{list})$ are the minimum and maximum values in the list, respectively.
\subsection{Correlation test on novelty score prediction with different section combinations}
\label{sec4.4}
\noindent We will answer RQ1 by examining the correlation between the prediction and the ground truth in this section. 
\subsubsection{Correlation analysis between prediction by PLMs and ground truth}
In addition to the traditional metrics of accuracy and absolute difference for the score prediction task, we also calculate Pearson, Spearman, and Kendall’s tau correlation coefficient. These three-correlation coefficient metrics are commonly used to evaluate the score prediction capabilities of models, as demonstrated in studies on using LLMs to evaluate abstractive summaries \citep{r39} and machine translations \citep{r40}. The result of correlation coefficient between PLMs prediction and ground truth is shown in Table \ref{tab4}.\\

\begin{landscape}
	\begin{table*}[h]
		\centering
		\caption{\centering{The result of the correlation coefficient between the PLMs’s prediction and the ground truth.}} 
		\label{tab4}
		\begin{tabular}{@{}m{10mm}<{\centering}m{10mm}<{\centering}m{10mm}<{\centering}m{10mm}<{\centering}m{10mm}<{\centering}m{10mm}<{\centering}m{10mm}<{\centering}m{10mm}<{\centering}m{10mm}<{\centering}m{10mm}<{\centering}m{10mm}<{\centering}m{10mm}<{\centering}m{10mm}<{\centering}m{10mm}<{\centering}m{10mm}<{\centering}m{10mm}<{\centering}@{}}
			\toprule
			\multirow{2}{*}{\textbf{SC}}&\multicolumn{3}{c}{\textbf{Longformer}} &\multicolumn{3}{c}{\textbf{BigBird}}&\multicolumn{3}{c}{\textbf{LongT5}}&\multicolumn{3}{c}{\textbf{LED}}&\multicolumn{3}{c}{\textbf{SciBERT}}\\
			&\textbf{P}&\textbf{SP}&\textbf{K}&\textbf{P}&\textbf{SP}&\textbf{K}&\textbf{P}&\textbf{SP}&\textbf{K}&\textbf{P}&\textbf{SP}&\textbf{K}&\textbf{P}&\textbf{SP}&\textbf{K}\\
			\midrule
			T	&0.0488	&0.0602	&0.0481	&0.0491	&0.0606	&0.0483	&0.0488	&0.0603	&0.0480	&0.0491	&0.0608	&0.0485	&0.0492	&0.0610	& 0.0488\\
			A	&0.0890	&0.0832	&0.0879	&0.0891	&0.0835	&0.0882	&0.0892	&0.0836	&0.0882	&0.0892	&0.0836	&0.0882	&0.0894	&0.0838	& 0.0884\\
			TA	&$0.1555^b$	&$0.1503^b$	&$0.1483^b$	&$0.1656^b$	&$0.1604^b$	&$0.1594^b$	&$0.1660^b$	&$0.1659^b$	&$0.1601^b$	&$0.1662^b$	&$0.1662^b$	&$0.1651^b$	&$0.1885^b$	&$0.1885^b$	& $0.1816^b$\\
			I	&$0.2002^c$	&$0.1976^c$	&$0.1951^c$	&$0.2021^c$	&$0.2019^c$	&$0.1975^c$	&$0.2011^c$	&$0.1990^c$	&$0.1964^c$	&$0.2020^c$	&$0.1999^c$	&$0.1970^c$	&$0.2030^c$	&$0.2010^c$	&$0.1983^c$\\
			IM	&$0.2150^c$	&$0.2105^c$	&$0.2077^c$	&$0.2180^c$	&$0.2134^c$	&$0.2110^c$	&$0.2161^c$	&$0.2115^c$	&$0.2080^c$	&$0.2161^c$	&$0.2115^c$	&$0.2090^c$	&$0.2199^c$	&$0.2140^c$	&$0.2122^c$\\
			IMR	&$0.2811^a$	&$0.2740^a$	&$0.2730^a$	&$0.2847^a$	&$0.2781^a$	&$0.2770^a$	&$0.2811^a$	&$0.2741^a$	&$0.2733^a$	&$0.2830^a$	&$0.2761^a$	&$0.2751^a$	&$0.2867^a$	&$0.2791^a$	&$0.2785^a$\\
			IMD	&$0.2830^b$	&$0.2772^b$	&$0.2741^b$	&$0.2850^b$	&$0.2781^b$	&$0.2773^b$	&$0.2859^b$	&$0.2801^b$	&$0.2777^b$	&$0.2850^b$	&$0.2783^b$	&$0.2777^b$	&$0.2881^b$	&$0.2810^b$	&$0.2804^b$\\
			IMRD	&$0.2711^c$	&$0.2750^b$	&$0.2695^b$	&$0.2721^c$	&$0.2771^b$	&$0.2710^b$	&$0.2710^c$	&$0.2755^b$	&$0.2688^b$	&$0.2730^c$	&$0.2771^b$	&$0.2710^b$	&$0.2850^c$	&$0.2889^b$	&$0.2831^b$\\
			IR	&$0.2170^c$	&$0.2131^c$	&$0.2100^c$	&$0.2200^c$	&$0.2165^c$	&$0.2131^c$	&$0.2183^c$	&$0.2140^c$	&$0.2111^c$	&$0.2194^c$	&$0.2150^c$	&$0.2122^c$	&$0.2222^c$	&$0.2181^c$	&$0.2155^c$\\
			IRD	&$\textbf{0.3181}^c$	&$\textbf{0.3080}^c$	&$\textbf{0.3045}^c$	&$\textbf{0.3205}^c$	&$\textbf{0.3104}^c$	&$\textbf{0.3060}^c$	&$\textbf{0.3211}^c$	&$\textbf{0.3110}^c$	&$\textbf{0.3071}^c$	&$\textbf{0.3201}^c$	&$\textbf{0.3103}^c$	&$\textbf{0.3061}^c$	&$\textbf{0.3261}^c$	&$\textbf{0.3154}^c$	&$\textbf{0.3125}^c$\\
			ID	&$0.2774^c$	&$0.2701^c$	&$0.2671^c$	&$0.2793^c$	&$0.2721^c$	&$0.2690^c$	&$0.2801^c$	&$0.2735^c$	&$0.2701^c$	&$0.2790^c$	&$0.2721^c$	&$0.2692^c$	&$0.2810^c$	&$0.2741^c$	&$0.2711^c$\\
			M	&$0.2090^c$	&$0.2021^c$	&$0.2000^c$	&$0.2112^c$	&$0.2041^c$	&$0.2020^c$	&$0.2101^c$	&$0.2030^c$	&$0.2010^c$	&$0.2091^c$	&$0.2020^c$	&$0.2000^c$	&$0.2130^c$	&$0.2065^c$	&$0.2043^c$\\
			MR	&0.1767	&0.1704	&0.1701	&0.1805	&0.1744	&0.1735	&0.1821	&0.1744	&0.1743	&0.1812	&0.1734	&0.1743	&0.1854	&0.1789	&0.1778\\
			MD	&$0.1681^a$	&0.1612	&0.1603	&$0.1689^a$	&0.1621	&0.1612	&$0.1601^a$	&0.1630	&0.1621	&$0.1688^a$	&0.1621	&0.1612	&$0.1754^a$	&0.1681	&0.1671\\
			MRD	&$0.1690^a$	&$0.1621^a$	&$0.1601^a$	&$0.1712^a$	&$0.1643^a$	&$0.1621^a$	&$0.1701^a$	&$0.1634^a$	&$0.1610^a$	&$0.1721^a$	&$0.1651^a$	&$0.1634^a$	&$0.1751^a$	&$0.1683^a$	&$0.1667^a$\\
			R	&$0.1931^c$	&$0.1761^c$	&$0.1743^a$	&$0.1951^c$	&$0.1783^c$	&$0.1761^a$	&$0.1985^c$	&$0.1818^c$	&$0.1793^b$	&$0.1961^c$	&$0.1792^c$	&$0.1774^b$	&$0.1991^c$	&$0.1834^c$	&$0.1812^b$\\
			RD	&$0.1312^a$	&$0.1081^a$	&$0.1067^a$	&$0.1333^a$	&$0.1102^a$	&$0.1091^a$	&$0.1321^a$	&$0.1092^a$	&$0.1083^a$	&$0.1333^a$	&$0.1110^a$	&$0.1101^a$	&$0.1362^a$	&$0.1132^a$	&$0.1110^a$\\
			D	&$0.1901^c$	&$0.1732^c$	&$0.1711^c$	&$0.1923^c$	&$0.1741^c$	&$0.1725^c$	&$0.1912^c$	&$0.1734^c$	&$0.1721^c$	&$0.1934^c$	&$0.1762^c$	&$0.1745^c$	&$0.1962^c$	&$0.1899^c$	&$0.1783^c$\\
			\bottomrule
		\end{tabular}
		\begin{tablenotes}
			\footnotesize
			\item \textbf{Note:} a represents p$<$0.05; b represents p$<$0.01; c represents p$<$0.001. SC, T, A, I, M, R, D respectively represent section combination, title, abstract, introduction, methods, results, and discussion. For example, TA means the model input consists of the title and abstract, with other letter combinations following the same pattern. SciBERT is Longformer version. P, SP and K is Pearson, Spearman and Kendall’s tau correlation coefficient, respectively. 
		\end{tablenotes}
	\end{table*}
\end{landscape}
\indent From the results in the Table \ref{tab4}, we can be observed that the predicted novelty scores based on all section combinations positively correlate with the actual scores. We attribute this to the fact that various sections of an academic paper, including the title and abstract, contain information that aids in predicting the novelty score. SciBERT achieved the highest correlation coefficient, which is closely related to its training corpus, enabling it to make relatively accurate predictions for novelty score prediction.  Furthermore, we can observe that the prediction and true score correlation coefficients are highest when using combinations of three-section structures as input, especially IRD or any combination that includes the introduction. This indicates that for the task of predicting novelty scores, combinations of three-section structures, particularly those including the introduction, are more effective. It is worth noting that while the correlation coefficient for IMRD is not the highest, it remains competitive with other high-accuracy combinations. This suggests that when the model uses IMRD as input, the information from various section structures can influence its judgment, leading to incorrect predictions. Although these predictions are inaccurate, the difference from the true scores is relatively small, resulting in lower accuracy but relatively higher correlation coefficients.
\subsubsection{Correlation analysis between prediction by LLMs and ground truth}
\noindent To further evaluate the ability of LLMs in predicting novelty scores, we calculated three correlation coefficients between the generated predictions and the real scores. The detailed results are presented in Table \ref{tab5}.\\
\begin{table*}[h]
	\centering
	\caption{\centering{The result of the correlation coefficient between the LLM's prediction and the ground truth.}} 
	\label{tab5}
	\begin{tabular}{@{}m{15mm}<{\centering}m{15mm}<{\centering}m{15mm}<{\centering}m{15mm}<{\centering}m{18mm}<{\centering}m{18mm}<{\centering}m{18mm}<{\centering}@{}}
		\toprule
		\multirow{2}{*}{\textbf{SC}}&\multicolumn{3}{c}{\textbf{GPT-3.5}} &\multicolumn{3}{c}{\textbf{GPT-4o}}\\
		&\textbf{P}&\textbf{SP}&\textbf{K}&\textbf{P}&\textbf{SP}&\textbf{K}\\
		\midrule
		T	&0.0248	&0.0250	&0.0227	&0.1035	&0.1035	&0.0976\\
		A	&0.1525	&0.1566	&0.1433	&-0.0295	&-0.1100	&-0.0277\\
		TA	&-0.1054	&-0.1054	&-0.0994	&0.1581	&0.1474	&0.1381\\
		I	&-0.0787	&-0.0565	&-0.0522	&0.0192	&0.0201	&0.0199\\
		IM	&-0.0674	&-0.0453	&-0.0416	&0.0199	&0.0318	&0.0299\\
		IMR	&\textbf{0.1550}	&0.1432	&0.1328	&$0.1916^a$	&$0.1916^a$	&$0.1806^a$\\
		IMD	&-0.1021	&-0.1243	&-0.1163	&0.1602	&0.1643	&0.1536\\
		IMRD	&-0.0815	&-0.1240	&-0.1160	&0.1617	&0.1789	&0.1681\\
		IR	&0.0386	&0.0420	&0.0386	&$0.1921^a$	&$0.1896^a$	&$0.1771^a$\\
		IRD	&-0.1129	&-0.1002	&-0.0936	&0.0207	&0.0207	&0.0195\\
		ID	&0.0404	&0.0721	&0.0670	&0.1038	&0.1038	&0.0979\\
		M	&-0.1517	&-0.1708	&-0.1566	&0.0545	&0.0803	&0.0744\\
		MR	&0.0809	&0.0502	&0.0456	&0.0793	&0.0808	&0.0755\\
		MD	&0.0354	&0.0499	&0.0462	&0.1090	&0.1088	&0.1003\\
		MRD	&-0.1066	&-0.1108	&-0.1019	&0.0779	&0.0803	&0.0755\\
		R	&0.1273	&\textbf{0.1603}	&\textbf{0.1510}	&0.1527	&0.1554	&0.1460\\
		RD	&0.0185	&0.0659	&0.0612	&\boldmath{$0.2389^a$}	&\boldmath{$0.2329^b$}	&\boldmath{$0.2186^a$}\\
		D	&0.0748	&0.0605	&0.0564	&0.0640	&0.0536	&0.0472\\
		\bottomrule
	\end{tabular}
	\begin{tablenotes}
		\footnotesize
		\item \textbf{Note:} a represents p$<$0.05; b represents p$<$0.01; c represents p$<$0.001. SC is Section Combination. P, SP and K is Pearson, Spearman and Kendall’s tau correlation coefficient, respectively. The other letter abbreviations are the same as Table \ref{tab4}.
	\end{tablenotes}
\end{table*}
\indent First, we analyzed the results of GPT-3.5. The results indicate that GPT-3.5's predictions for many section combinations exhibit a negative correlation with the real scores. This suggests that GPT-3.5 is not adept at the task of novelty score prediction, likely due to the hallucination problem inherent in LLMs. For the results generated by GPT-3.5, we need to further process many additional contents generated that is consistent with our requirements. This may also be the reason for the negative correlation in the correlation coefficient calculation results of GPT-3.5. Furthermore, we observe that the correlation coefficients for the IMR and R section combinations are relatively high. This indicates that using a combination of three-section structures is relatively suitable for predicting novelty scores.\\
\indent Then, we analyze the results of GPT-4o. From the correlation coefficient calculations, we see that nearly all results exhibit a positive correlation. This indicates that GPT-4o is capable of understanding and performing the task to a certain extent. Compared to the accuracy and F1 scores, the correlation coefficient results for GPT-4o predictions are more impressive. Unlike the PLMs, the section combinations that include the Results section show higher correlation coefficients. We believe this indicates that GPT-4o places greater emphasis on the impact of the Results section on novelty scores. Additionally, we think the Discussion section overlaps with the Introduction, which may explain why the IR and RD combinations outperform others. The poorer performance of the IRD combination is likely due to the confusion caused by the overlap between the Introduction and Discussion sections. Additionally, we can observe that the IMR and IMD combinations also exhibit competitive results, indicating that the Methods section holds considerable significance as well.

\subsection{Novelty score prediction performance with different section combinations}
\label{subsec4.1}
\noindent We answer RQ2 by analyzing the performance of the PLMs and LLMs in the novelty score prediction task in this section.
\begin{landscape}
\begin{table*}[h]
	\centering
	\caption{\centering{The results of different section combinations of PLMs in novelty score prediction tasks.}} 
	\label{tab2}
	\begin{tabular}{@{}m{15mm}<{\centering}m{15mm}<{\centering}m{15mm}<{\centering}m{15mm}<{\centering}m{15mm}<{\centering}m{15mm}<{\centering}m{15mm}<{\centering}m{15mm}<{\centering}m{15mm}<{\centering}m{15mm}<{\centering}m{15mm}<{\centering}@{}}
		\toprule
		\multirow{2}{*}{\textbf{SC}}&\multicolumn{2}{c}{\textbf{Longformer}} &\multicolumn{2}{c}{\textbf{BigBird}}&\multicolumn{2}{c}{\textbf{LongT5}}&\multicolumn{2}{c}{\textbf{LED}}&\multicolumn{2}{c}{\textbf{SciBERT}}\\
		&\textbf{Acc}&\bm{$F_1$}&\textbf{Acc}&\bm{$F_1$}&\textbf{Acc}&\bm{$F_1$}&\textbf{Acc}&\bm{$F_1$}&\textbf{Acc}&\bm{$F_1$}\\
		\midrule
		T & 0.5274&0.5168&0.5265&0.5157&0.5266&0.5159&0.5281&0.5172&0.5283&0.5175\\
		A & 0.5331&0.5312&0.5334&0.5315&0.5332&0.5314&0.5335&0.5318&0.5338&0.5320\\
		TA & 0.5943&0.5880&0.5971&0.5900&0.5951&0.5891&0.5940&0.5891&0.5982&0.5901\\
		I & 0.6114&0.5863&0.6082&0.5811&0.6102&0.5843&0.6123&0.5861&0.6122&0.5854\\
		IM&0.6114&0.5903&0.6130&0.5921&0.6122&0.5911&0.6112&0.5891&0.6152&0.5922\\
		IMR&0.6577&0.6351&0.6550&0.6309&0.6574&0.6352&0.6562&0.6354&0.6600&0.6381\\
		IMD&0.6572&0.6281&0.6541&0.6322&0.6567&0.6281&0.6573&0.6543&0.6592&0.6314\\
		IMRD&0.6145&0.5940&0.6101&0.592&0.6167&0.5921&0.6143&0.5952&0.6182&0.5973\\
		IR&0.6200&0.6143&0.6222&0.6154&0.6200&0.6152&0.6201&0.6152&0.6221&0.6152\\
		IRD&\textbf{0.6681}&\textbf{0.6455}&\textbf{0.6701}&\textbf{0.6503}&\textbf{0.6692}&\textbf{0.6481}&\textbf{0.6778}&\textbf{0.6462}&\textbf{0.6824}&\textbf{0.6515}\\
		ID&0.6515&0.6333&0.6534&0.6341&0.6563&0.6333&0.6512&0.6322&0.6532&0.6354\\
		M&0.6000&0.5781&0.6010&0.5795&0.6021&0.5794&0.6014&0.5781&0.6031&0.5812\\
		MR&0.5801&0.5601&0.5834&0.5622&0.5841&0.5643&0.5812&0.5589&0.5854&0.5632\\
		MD&0.5861&0.5757&0.5861&0.5765&0.5885&0.5785&0.5867&0.5750&0.5902&0.5824\\
		MRD&0.5931&0.5862&0.5951&0.5881&0.5961&0.5892&0.5900&0.5881&0.5973&0.5996\\
		R&0.5820&0.5670&0.5834&0.5681&0.5887&0.5733&0.5824&0.5671&0.5912&0.5793\\
		RD&0.5500&0.5341&0.5523&0.5364&0.5601&0.5410&0.5559&0.5342&0.5697&0.5512\\
		D&0.5689&0.5650&0.5712&0.5662&0.5750&0.5681&0.5767&0.5681&0.5813&0.5667\\
		\bottomrule
	\end{tabular}
	\begin{tablenotes}
		\footnotesize
		\item \textbf{Note:} SC, T, A, I, M, R, D respectively represent section combination, title, abstract, introduction, methods, results, and discussion. For example, TA means the model input consists of the title and abstract, with other letter combinations following the same pattern. SciBERT is Longformer version. $F_1$ is $Weighted\_F1$.
	\end{tablenotes}
\end{table*}
\end{landscape}

\subsubsection{Results of PLMs with different section combinations}
\noindent To address this research question, we fine-tuned five PLMs on the novelty score prediction task using different section combinations. The training process spanned 10 epochs with an initial learning rate of 0.0001, which gradually decreased. To prevent overfitting, we implemented an early stopping strategy, terminating training if the F1 score on the validation set did not improve for three consecutive epochs. The final results are shown in Table \ref{tab2}.\\
\indent As shown in Table \ref{tab2}, the performance of the five PLMs on the novelty score prediction task does not differ significantly. However, SciBERT slightly outperforms the other models, likely because it is trained on scientific-related corpora. While the other pre-trained models also include academic papers in their training data, their performance may be affected by the presence of other non-academic corpora. Furthermore, the results show that input combinations including the introduction tend to perform better than other combinations. We attribute this to the fact that authors typically articulate the contributions or innovative aspects of their papers in the introduction, which are relevant to assessing novelty. These details likely influence the model's judgment of the novelty score. Moreover, considering the performance of each model, input combinations with three sections yield the highest accuracy and F1 scores, with the IRD combination performing the best, followed by IMR and IMD. We believe that while the introduction already contains significant information relevant to predicting novelty scores , such as the main contributions proposed by the author, the inclusion of additional sections further enhances the model's understanding of the task, leading to improved results. Finally, we observe that model performance of IMRD is not good. We think that the content and knowledge encompassed in the full text of a paper are too extensive for current PLMs to capture knowledge effectively for the task of novelty score prediction. Furthermore, we observed that the highest accuracy achieved was only 0.682. Therefore, we conclude that the task of novelty score prediction poses a significant challenge for PLMs.
\begin{table*}[h]
	\centering
	\caption{\centering{The results of LLMs on the task of novelty score prediction.}} 
	\label{tab3}
	\begin{tabular}{@{}m{40mm}<{\centering}m{20mm}<{\centering}m{20mm}<{\centering}m{20mm}<{\centering}m{20mm}<{\centering}@{}}
		\toprule
		\multirow{2}{*}{\textbf{Section Combination}}&\multicolumn{2}{c}{\textbf{GPT-3.5}} &\multicolumn{2}{c}{\textbf{GPT-4o}}\\
		&\textbf{Acc}&\bm{$F_1$}&\textbf{Acc}&\bm{$F_1$}\\
		\midrule
		T&0.3333	&0.2508	&0.3083	&0.2418\\
		A&0.3500	&0.2836	&0.3333	&0.2145\\
		TA	&0.3500	&0.2803	&0.3083	&0.2230\\
		I	&0.2917	&0.2220	&0.3000	&0.2493\\
		IM	&0.3167	&0.2676	&0.3500	&0.2800\\
		IMR	&0.3917	&\textbf{0.3388}&0.3417	&0.2616\\
		IMD	&0.2833	&0.2260	&0.3167	&0.2471\\
		IMRD	&0.3083	&0.2841	&\textbf{0.3917}	&0.3088\\
		IR	&\textbf{0.4000}	&0.3194	&0.3833	&0.3161\\
		IRD	&0.2583	&0.1813	&0.2917	&0.2296\\
		ID	&0.3917	&0.3097	&0.3750	&0.2970\\
		M	&0.2667	&0.2351	&0.3333	&0.2744\\
		MR	&0.3417	&0.3094	&0.3000	&0.2420\\
		MD	&0.3250	&0.2701	&0.3333	&0.2902\\
		MRD	&0.3167	&0.2587	&0.3333	&0.2786\\
		R	&0.3417	&0.2815	&0.3750	&0.2912\\
		RD	&0.3250	&0.2457	&0.3833	&\textbf{0.3241}\\
		D	&0.3500	&0.3018	&0.2833	&0.2704\\
		\bottomrule
	\end{tabular}
	\begin{tablenotes}
		\footnotesize
		\item \textbf{Note:} The letter abbreviations are the same as Table \ref{tab2}. $F_1$ is $Weighted\_F1$.
	\end{tablenotes}
\end{table*}
\subsubsection{Result of novelty score prediction using LLM}
\noindent We tested the performance of two LLMs (GPT-3.5 and GPT-4o) on the task of predicting novelty scores based on different section combinations. All results were obtained in the form of zero shot. To mitigate the randomness of the generated results, we conducted five predictions and selected the most frequently occurring score. The results are presented in Table \ref{tab3}. \\
\indent From the results, we can see that the novel score prediction task also poses challenges for these two LLMs, with the highest accuracy only reaching 0.4. We attribute the suboptimal performance of the LLMs to several factors. Firstly, in a zero-shot scenario, LLMs struggle to grasp the patterns associated with novelty score prediction, leading to erroneous judgments. Secondly, the training corpus of these models may not include data relevant to the novelty score task. Thirdly, the models' limited reasoning capabilities hinder their ability to infer the task content based on the provided prompts. From the perspective of different section combinations, the results for both GPT-3.5 and GPT-4o are quite unstable. GPT-3.5 performs better when the input is IR, while GPT-4o performs better with IMRD as the input. We believe this instability is related to the tendency of LLMs to generate favorable responses, as they are inclined to provide satisfactory answers or predictions. Finally, for the results generated from LLM, we examined the generated content and found that LLM tends to give the highest novelty score (highly novel). We think this tendency contributes to the low accuracy and F1 scores.
\begin{table*}[h]
	\centering
	\caption{\centering{The result of traditional method and best performance by our method.}} \label{tab6}
	\begin{tabular}{@{}m{30mm}<{\centering}m{15mm}<{\centering}m{15mm}<{\centering}m{15mm}<{\centering}m{15mm}<{\centering}m{15mm}<{\centering}@{}}
		\toprule
		\textbf{Method}&\textbf{Acc}&\bm{$F_1$}&\textbf{P}&\textbf{SP}&\textbf{K}\\
		\midrule
		Shibayama's method&0.4265&0.3637&0.0132&0.0362&0.0346\\
		SciBERT+IRD&0.6824&0.6515&$0.3261^c$&$0.3154^c$&$0.3125^c$\\
		\bottomrule
	\end{tabular}
	\begin{tablenotes}
		\footnotesize
		\item \textbf{Note: a represents p$<$0.05; b represents p$<$0.01; c represents p$<$0.001. The letter abbreviations are the same as Table \ref{tab2}. $F_1$ is $Weighted\_F1$}.
	\end{tablenotes}
\end{table*}	
\subsection{Comparison with a traditional novelty evaluation method}
In this section, we compare the best-performing section combinations (SciBERT+IRD) with a traditional novelty evaluation method mentioned in Section \ref{sec4.2}. The results are shown in Table \ref{tab6}. As we can observe, our method outperforms the traditional method in terms of both accuracy and $F_1$ score, as well as the results of the correlation calculations. This demonstrates that using comprehensive text information to assess a paper's novelty is more effective than the traditional citation-based methods. Furthermore, the results indicate that the comprehensive text information indeed contains more content that contributes to novelty evaluation, which is also supported by the findings in Sections \ref{sec4.4} and \ref{subsec4.1}. This further underscores the potential of leveraging a broader range of textual data for novelty detection. 
\subsection{Section association analysis for automated paper novelty assessment}
\noindent In this section, we will comprehensively analyze the results of PLMs and LLMs to answer RQ3.\\
\indent Based on the results from the first two research questions, we can proceed to discuss the third research question. As seen in the results from Table \ref{tab2} and Table \ref{tab4}, models that include the introduction section consistently perform better than other combinations. Therefore, we think the introduction to be a crucial section for evaluating the novelty of a paper. Furthermore, we observe that the combination of the introduction and results sections also achieves notable performance, indicating that this combination is beneficial for evaluating the novelty of a paper. Although these combinations did not perform well with the LLMs, Tables \ref{tab3} and \ref{tab5} show that inputs including the results section perform relatively well. Therefore, we also consider the results section to be essential for assessing the novelty of a paper. Furthermore, we observe that the combination of the introduction, results, and discussion sections yields the best performance with the PLMs. Similarly, the combination of results and discussion also performs well with the LLMs. Based on this, we can infer that the discussion section serves as a valuable supplement to the assessment of a paper's novelty, building on the information provided in the introduction and results sections. Although the methods section is a crucial section for evaluating a paper's novelty, its text description is often abstract, and the content related to novelty can be challenging for machines to comprehend. Therefore, it may not be suitable for automated assessment of a paper's novelty. We believe that evaluating the novelty of the methods section requires external knowledge for support, such as the opinions of peer review experts. In summary, we believe that the combination of the introduction, results, and discussion sections is crucial for the automated assessment of a paper's novelty.

\begin{figure*}[p]%
	\centering
	\includegraphics[width=1.2\textwidth]{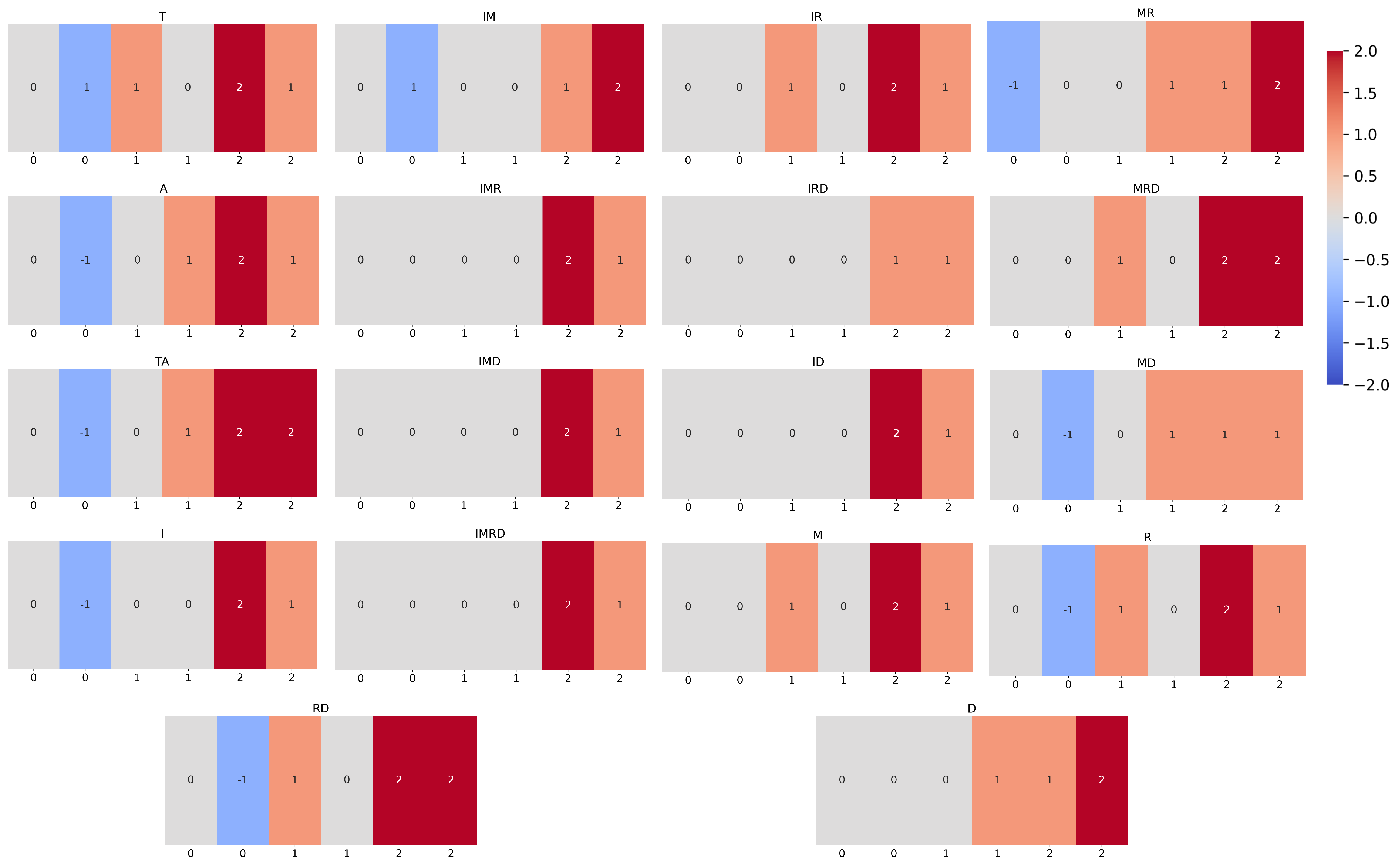}
	\vspace{-0.5cm}
	\begin{flushleft}
		{\footnotesize \textbf{Note}: The x axis represents the true labels, and the numbers in different colors within each plot indicate the difference between the prediction and true labels. The legend is on the right, with darker colors indicating a greater difference between the predicted labels and the true labels.} 
	\end{flushleft}
	\caption{\centering{Case study on PLM (Longformer version of SciBERT) for novelty score prediction using different paper sections as input.}}
	\label{fig:6}
\end{figure*}

\subsection{Case study}
\begin{figure*}[p]%
	\centering
	\includegraphics[width=1.2\textwidth]{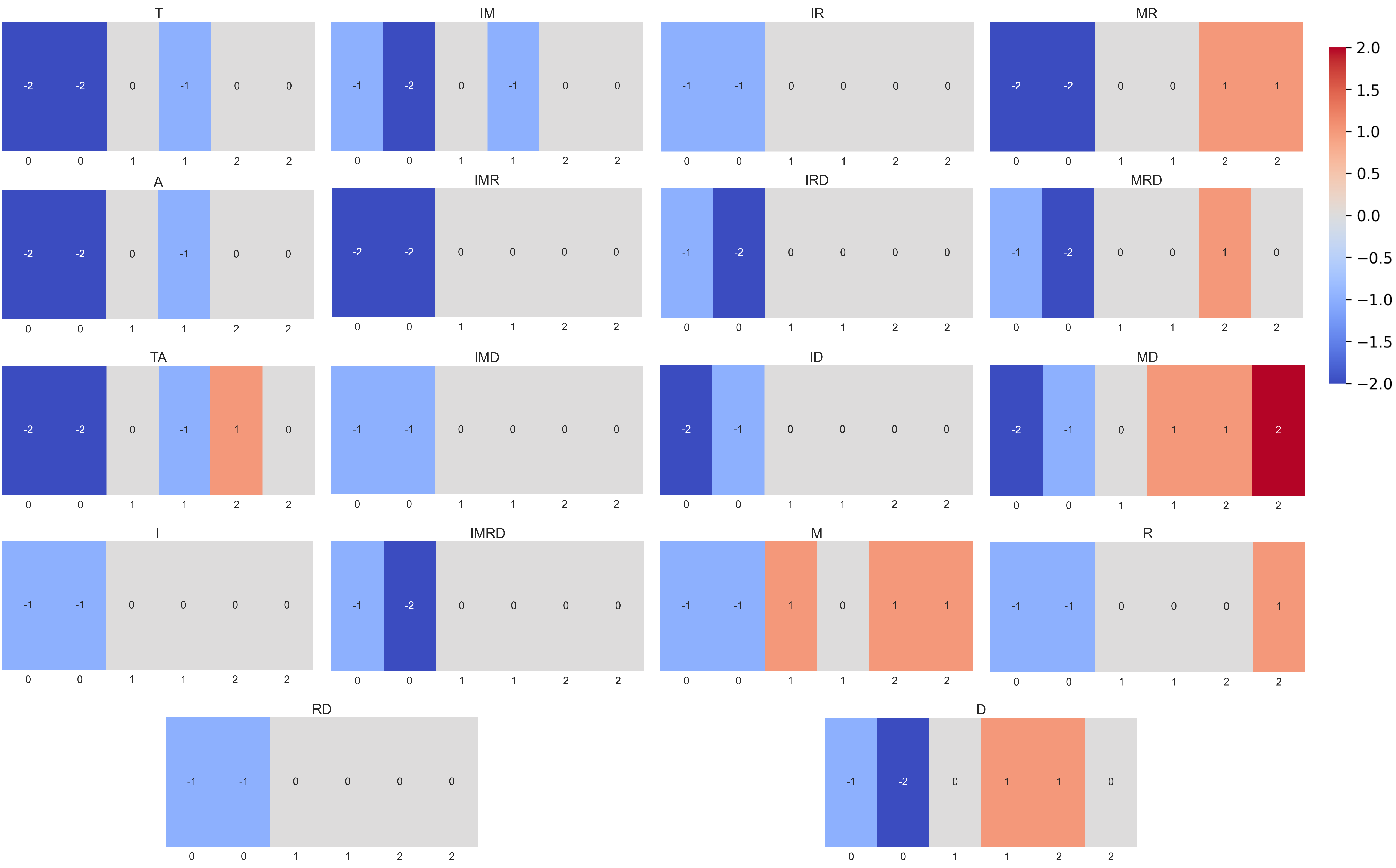}
	\vspace{-0.9cm}
	\begin{flushleft}
		{\footnotesize \textbf{Note}: The x axis represents the true labels, and the numbers in different colors within each plot indicate the difference between the prediction and true labels. The legend is on the right, with darker colors indicating a greater difference between the predicted labels and the true labels.} 
	\end{flushleft}
	\caption{\centering{Case study on LLM (GPT-4o) for novelty score prediction using different paper sections as input.}}
	\label{fig:7}
\end{figure*}

\noindent As shown in Figure \ref{fig:6} and \ref{fig:7}, we conducted case studies on PLM (Longformer version of SciBERT) and LLM (GPT-4o) based on different section combinations inputs. We treat the model's output and the ground truth as two rows in a matrix, with the top row representing the labels and the bottom row representing the predictions. The letter combinations above each matrix represent the abbreviations for different section structures: I stand for Introduction, M for Methods, R for Results, D for Discussion, T for Title, and A for Abstract. For example, IM represents the combination of Introduction and Methods. When the true label is 0, the range of discrepancies is $[-2,0]$; when the true label is 1, the range is $[-1,1]$; and when the true label is 2, the range is $[0,2]$. For example, if the true label is 0 and the prediction is 2, the discrepancy is -2, resulting in a darker color in the plot.\\

\indent We randomly selected two papers from each of the three novelty score categories, making a total of six papers for case studies. First, we examined the predictions of the PLM under different section combinations. From the Figure \ref{fig:6}, it is evident that the PLM struggles to accurately predict high novelty scores but can generally make correct predictions for medium and low novelty scores. Regarding the proximity of score predictions, section combinations that include the Introduction achieve better results. Combinations of three sections contain Introduction perform better than others, particularly the IRD combination. Despite predicting high novelty scores incorrectly, it still provides a proximate score of 2.\\
\indent Secondly, we conducted an examination of LLM in Figure \ref{fig:7}, the prediction results indicate that the LLM is proficient in judging papers with novelty scores of 1 and 2. However, for many papers with a novelty score of 0, the LLM often predicts a score of 2. This suggests that the LLM tends to provide favorable results and struggles to identify papers with low novelty. Considering different chapter structures, the predictions for I, IMD, and RD are the closest to the ground truth. Although the number of well-performing section combinations varies, it is evident that the introduction, results, and discussion sections are beneficial for predicting novelty scores. Furthermore, the inclusion of the methods section, supplemented by the introduction and discussion, allows the LLM to better understand the paper's content and make more accurate novelty score predictions.\\
\indent In summary, through case studies involving PLM and LLM, we can believe that the combination of introduction, results, and discussion sections is more effective for novelty score prediction.
\section{Discussion}
\noindent In this section, we will discuss the implication of our study on theoretical and practical, and limitation of our study.
\subsection{Implication}
\subsubsection{Theoretical Implication}
\noindent In this study, we collected all the PDF data of papers and their corresponding peer review reports from ICLR 2022 and 2023. We then parsed the PDFs and used deep learning models to identify the main text of the papers. Subsequently, we employed deep learning models and LLMs to recognize the structure of the identified main text. Then, we used the novelty score given by the reviewers in the review report as the standard, and fine-tuned the PLMs with different section combinations as inputs. We also verified the performance of PLMs with different section combinations as inputs. Additionally, we conducted small-sample tests to assess the performance of LLMs when provided with different section combinations as content prompts.\\
\indent Based on the results we obtained, both PLMs and LLMs can, to some extent, combine section structure information to predict novelty scores in long-text processing; however, each has its limitations. First, for PLMs, although their accuracy and $F_1$ scores are relatively high, they appear to be less sensitive to papers with high novelty scores and fail to make accurate predictions in such cases. We believe that the relatively small number of papers with high novelty scores is one reason for this phenomenon. Regarding LLMs, although their performance across various metrics is not ideal, they seem to perform better in predicting papers with high novelty scores. We speculate that this may be because LLMs, which are primarily designed for conversational tasks, tend to exhibit a more favorable and engaging tone. Based on this, we argue that relying solely on PLMs or LLMs for automatic novelty score prediction could lead to polarized results. Since both PLMs and LLMs have their respective strengths, we believe that combining the advantages of both through effective methods may provide a more optimal solution for automatic novelty score prediction.\\
\indent Finally, our results indicate that the optimal section combination for predicting novelty scores is the introduction, results, and discussion. We believe that the introduction section contains the main contributions and innovations proposed by the authors, which are crucial for assessing novelty. The results and discussion sections provide specific explanations of these contributions and further summarize the findings, enabling the model to better understand the task of predicting novelty scores and thus achieve the best performance. While our findings suggest that certain section combinations may enhance classifier performance, these results should not be interpreted as a definitive measure of novelty. Instead, they highlight text-based features that may signal novel contributions, which require further validation through human judgment. Additionally, the method in this study holds theoretical value for fields beyond computer science, particularly for papers in domains where section structure Identification is possible. It is applicable if the content of the paper can be classified into the IMRaD structure, enabling novelty score prediction, or by labeling enough novelty scores for transfer learning.
\subsubsection{Practical Implication}
\noindent Firstly, we provide insights for future evaluations of paper novelty. When assessing novelty using the core content of a paper rather than partial elements (such as abstracts, keywords, etc.), priority should be given to the content found in the introduction, results, and discussion sections, particularly focusing on significant sentences and paragraphs. Additionally, if only a portion of the content, such as a single section, is to be considered, our results suggest a prioritization hierarchy. The introduction should be the primary focus, followed by the results and discussion sections.\\
\indent Secondly, in a zero-shot setting, LLMs are unreliable for the task of novelty score prediction. This is consistent with the findings of Mike Thelwall \citep{r27} and Zhou et al. \citep{r13}, who also concluded that current LLMs are insufficient for fine-grained evaluations. Overall, LLMs consistently generate results that tend to be overly satisfactory, often assigning high scores when assessing the novelty of papers. This indicates that the current LLM is not accurate enough to complete formal peer review related work, especially novelty assessment. We believe that fine-tuning these models could potentially enable them to assist humans in the review process.\\
\indent Thirdly, our results can provide guidance for novice reviewers. Young reviewers, who may be unsure how to assess the novelty of a paper in their initial reviews, can begin by focusing on the introduction, results, and discussion sections for a preliminary evaluation. Furthermore, they can leverage their accumulated knowledge to make more comprehensive judgments, thereby completing the review process with a clear and systematic approach.\\
\indent Finally, our research also can generalizable to other domains. Using ICLR papers and novelty scores as a reference domain, we explored the optimal combination of section structures for predicting novelty scores. To apply this to other domains, it is only necessary to identify the section structures in the target domain and provide the corresponding novelty scores. Transfer learning and other cross-domain learning methods can then be employed to select the optimal structure based on our conclusions. In addition to peer review in ICLR and the field of computer science, peer review in other domains also requires reviewers to assess the novelty of a paper. While this assessment may not always take the form of specific scores, it often involves qualitative grading akin to scoring. Therefore, our research can applicable to other fields as well.

\subsection{Limitation}
\noindent Our study has several limitations that warrant emphasis. First, We acknowledge that novelty assessment is a multifaceted process influenced by factors such as the reviewer’s background, familiarity with cutting-edge research, and interpretation of original contributions. While our method focuses on text analysis, it does not claim to capture the full complexity of human judgment involved in novelty evaluation. The insight proposed in this study should be viewed as part of a broader toolkit for assessing novelty. By identifying patterns within chapter combinations and their potential influence on the perception of novelty, we aim to complement rather than replace the holistic evaluations conducted by reviewers. Additionally, we only tested two LLM for their performance in novelty score prediction;  many other available LLMs, such as LLaMA 3, remain to be tested. Furthermore, the prompts used in our study represent just one possible form and could benefit from further exploration to design more effective prompts. Although GPT -3.5 or GPT-4o currently possess certain capabilities, they also have limitations. GPT-3.5, for instance, still requires certain functionalities or additional tasks, such as incorporating current scientific databases, to be effectively utilized in these experiments. Our study did not account for the impact of other content in academic papers, such as tables, figures, and charts, which are crucial components of academic papers. Lastly, the data we used from ICLR is limited to the field of computer science, introducing certain constraints on the generalizability of our findings. \\
\indent Despite these limitations, this study offers a novel perspective on how text-based features can be leveraged to detect potential signals of novelty in academic writing. Our approach lays the groundwork for further exploration of the relationship between textual content and perceived innovation in research.
\section{Conclusion and future works}
\noindent In this study, we explored three research questions. To achieve this, we conducted section structure recognition on academic papers and fine-tuned the PLMs using different section combinations as inputs to verify its effectiveness in novelty score prediction tasks. Additionally, we evaluated the performance of LLMs on the same task. Finally, we discussed which section structures should be prioritized when assessing the novelty of academic papers. From the final results, the performance of the PLMs was generally moderate, with the highest accuracy for the section combination of Introduction, Results, and Discussion reaching only 0.682. Although this accuracy is not high, it still provides some insights. The LLMs performed even worse on the novelty score prediction task but showed decent results in terms of correlation coefficients. Our findings suggest that the Introduction, Results, and Discussion sections should be the primary focus when evaluating the novelty of academic papers.\\
\indent In the future, we plan to collect data from a wider array of disciplines and journals to assess novelty, as our current dataset is limited to the field of computer science and conference papers. We will design more effective model architectures to accomplish the task of novelty score prediction, while also incorporating traditional novelty evaluation methods may be achieve better results. More information, such as abstracts and keywords, can be included in the model to examine the effectiveness of these elements. Furthermore, we plan to develop strategies to unlock the potential of LLMs, enabling them to perform novelty score prediction more effectively. Exploring the integration of deep learning with LLMs is another area we intend to investigate. Moreover, future work could integrate additional dimensions of novelty assessment, such as reviewer expertise, citation patterns, and qualitative insights, to create a more comprehensive evaluation framework. 


\section{Acknowledgment}
\noindent This study is supported by the National Natural Science Foundation of China (Grant No. 72074113).

\bibliographystyle{elsarticle-harv} 
\bibliography{ref}

\end{document}